\documentclass[runningheads]{llncs}
\usepackage{graphicx}
\usepackage{caption}
\usepackage{booktabs}
\usepackage{ amssymb }
\usepackage{bbold}
\usepackage{gensymb}
\usepackage{multirow}
\usepackage{multicol}
\usepackage{marvosym}
\usepackage[colorlinks,linkcolor=blue]{hyperref}
\begin{document}
\title{Positional Contrastive Learning for Volumetric Medical Image Segmentation}
\titlerunning{Positional Contrastive Learning}

\author{Dewen Zeng\inst{1}$^{(\textrm{\Letter})}$ \and
Yawen Wu\inst{2} \and
Xinrong Hu\inst{1} \and
Xiaowei Xu\inst{3} \and 
Haiyun Yuan\inst{3} \and 
Meiping Huang\inst{3} \and 
Jian Zhuang\inst{3} \and 
Jingtong Hu\inst{2} \and 
Yiyu Shi\inst{1}$^{(\textrm{\Letter})}$
}
% index{Zeng, Dewen}
% index{Wu, Yawen}
% index{Hu, Xinrong}
% index{Xu, Xiaowei}
% index{Yuan, Haiyun}
% index{Huang, Meiping}
% index{Zhuang, Jian}
% index{Hu, Jingtong}
% index{Shi, Yiyu}
\authorrunning{D. Zeng, et al.}
%

% \authorrunning{blind review}
\institute{
University of Notre Dame, Notre Dame, IN, USA \\\email{\{dzeng2, yshi4\}@nd.edu}\and
University of Pittsburgh, Pittsburgh, PA, USA \and
Guangdong Provincial People's Hospital, Guangzhou, China
}
\maketitle              % typeset the header of the contribution
\begin{abstract}
The success of deep learning heavily depends on the availability of 
large labeled training sets.
% Ideally, if the training set covers all possibles cases in the real world application, DNNs can achieve extremely high accuracy.
However, it is hard to get large labeled datasets in medical image domain because of the strict privacy concern and costly labeling efforts. 
Contrastive learning, an unsupervised learning technique,   
has been proved powerful in learning image-level representations from unlabeled data.
The learned encoder can then be transferred or fine-tuned to improve the performance of downstream tasks with limited labels. 
A critical step in contrastive learning is the generation of contrastive data pairs, which is relatively simple for natural image classification but quite challenging for medical image segmentation due to the existence of the same tissue or organ across the dataset.
As a result, when applied to medical image segmentation, most state-of-the-art contrastive learning frameworks inevitably introduce a lot of false negative pairs and result in degraded segmentation quality. 
%Most existing contrastive learning frameworks target natural image classification, and use data augmentation such as flipping or rotating, which are not suitable for medical image segmentation  The pioneer work applying contrastive learning to medical image segmentation uses partition of 3D volumes and may still induce false negative pairs.     
To address this issue, we propose a novel positional contrastive learning (PCL) framework to generate contrastive data pairs by leveraging the position information in volumetric medical images.
Experimental results on CT and MRI datasets demonstrate that the proposed PCL method can substantially improve the segmentation performance compared to existing methods in both semi-supervised setting and transfer learning setting. \footnote[1]{Code available at \href{https://github.com/dewenzeng/positional_cl}{github.com/dewenzeng/positional\_cl}}

% \keywords{Contrastive Learning \and Self Supervised Learning \and Image Segmentation \and Deep Neural Network.}
\end{abstract}

\section{Introduction}
Deep neural networks (DNNs) play an important role in today's medical image segmentation \cite{ronneberger2015u,xu2019whole,wang2020ica,wang2019msu,isensee2018nnu}.
To achieve state-of-the-art accuracy, most of the existing methods rely on supervised learning when large labeled datasets can be used for training.
% However, due to the strict privacy concern and extensive annotation effort in the medical domain, acquiring such large labeled datasets is usually prohibitive.
However, due to the extensive annotation effort and the requirement of expertise in the medical domain, acquiring such large labeled datasets is usually prohibitive.
In the meantime, a large amount of unlabeled image data from modalities such as Computed Tomography (CT) and Magnetic Resonance Imaging (MRI) is generated every day all around the world.
Therefore, it is desirable that the DNNs can leverage the numerous unlabeled data to achieve higher performance with limited annotations.
Contrastive learning \cite{chen2020simple,chen2020big,he2020momentum,chen2020improved,misra2020self}, as a self-supervised learning (SSL) method, has shown great success in learning image-level features from large-scale unlabeled data without using any human-annotated labels.
The main idea of contrastive learning is to contrast the similarities of sample pairs in the representation space through contrastive loss, pulling the representations of similar pairs (a.k.a. positive pairs) together and pushing the representations of dissimilar pairs (a.k.a. negative pairs) apart.
In SSL setting, an encoder is trained using contrastive loss with unlabeled data.
After that, the trained encoder can be used as the initialization for training a supervised downstream task such as object detection and image segmentation.
Many works have shown that the encoder learned by contrastive learning performs better than the encoder trained with supervised learning \cite{he2020momentum,chen2020simple}.

Most existing contrastive learning 
frameworks are for image classification where the instances in two different images have dissimilar features. When directly applying them to medical image segmentation where different images can have similar structures or organs, a large number of false negative pairs will be induced, leading to degraded performance. 
Recently, \cite{chaitanya2020contrastive} attempted to address this issue through a global contrastive learning approach for 3D medical image segmentation.
It divides each volume into several partitions and considers the slices of corresponding partitions in different volumes as positive pairs and those of different partitions as negative pairs.
However, the last a few slices of a partition can be very similar to the first a few slices of the next partition as they are naturally adjacent, which may still result in many false negative pairs. 
% In addition, the scheme can still possibly pick up task-irrelevant features. 

To alleviate the problem, we propose a novel positional contrastive learning (PCL) framework, 
which generates contrastive data pairs based on the position of a slice in volumetric medical images. Slices that are close are considered positive pairs while those that are far apart are considered negative. Such a strategy can better leverage the domain-specific cue of medical images as adjacent slices typically contain similar anatomical structures, thus reducing false negatives. We evaluate the proposed PCL framework on two CT datasets and two MRI datasets. The experimental results show that our method can achieve better performance compared with state-of-the-art baselines in both semi-supervised and transfer learning settings.

\section{Related Work}
Recent years have seen powerful self-supervised visual feature learning approaches with DNNs.
By exploiting the information in large unlabeled datasets, a network can learn hierarchical features that can help the training of other downstream tasks, especially when the training labels of these tasks are limited. Early SSL methods are mostly based on the design of pretext tasks, in which pseudo labels are automatically generated for network training. As these methods rely on ad-hoc heuristics, the learned representation lack generality \cite{chen2020simple}.
Contrastive learning has recently become a prevailing SSL method because of its superior performance. In contrastive learning, 
a contrastive loss \cite{hadsell2006dimensionality} is used to enforce representations of positive pairs to be similar and those of negative pairs to be dissimilar \cite{he2020momentum,chen2020simple,misra2020self,tian2019contrastive,jiao2020self,li2020contrastive}.
MoCo \cite{he2020momentum} and SimCLR \cite{chen2020simple} are two different contrastive learning frameworks that yield state-of-the-art results.
MoCo maintains a dictionary as a queue to store negative samples for training, while SimCLR explores the use of in-batch samples for negative sampling.
% However, these approaches are based on instance discrimination for natural images.
% Directly applying them to medical image segmentation tasks will induce some problems. 
Most of these works are based on image classification tasks, assuming that the instances in two different images have dissimilar features. This is not the case, however, for medical images, because the same target organ or structure usually exists in all the images across the dataset. 
For example, in ACDC MICCAI 2017 dataset \cite{bernard2018deep}, the target structures such as the left ventricle and the right ventricle appear in almost every slice of the volumetric image for all patients. As such, if we follow the method used in image classification tasks and treat the augmented images from different slices as negative, many of them will actually be false negatives. 
% In addition, the difference between different images may contain task-irrelevant structures, which will lead to the learning of useless information and accordingly performance degradation. For example, the slices of the same patient in ACDC shows different lung structures, while lung is not among the target segmentation structures. Still, existing contrastive learning schemes will likely pick up features related to the lung, which would not help downstream segmentation.
% because there exist structural similarity and dissimilarity across images of the same patient and of different patients, simply using data augmentation to generate contrastive data pairs may result in learning task irrelevant features (e.g., irrelevant organs).

%For example, \cite{chen2020simple} learned an encoder on ImageNet without any label and achieved significant improvement in accuracy when the encoder was finetuned on other tasks such as image classification on CIFAR 10.

%\textbf{Pretext tasks.} 
%Various pretext tasks have been proposed and applied for SSL, examples include generation-based methods like image colorization \cite{zhang2016colorful} and image inpainting \cite{pathak2016context}, and context-based methods like jigsaw puzzle \cite{noroozi2016unsupervised,ahsan2019video} and geometric transformation \cite{gidaris2018unsupervised,xu2019self,chen2019self}.
% image clustering \cite{caron2018deep,caron2019unsupervised},

%\textbf{Contrative learning.} 

% Our contrastive framework is based on SimCLR \cite{chen2020simple} because of its simplicity and high performance.
The state-of-the-art contrastive learning method for medical image segmentation \cite{chaitanya2020contrastive} attempted to address this issue through the partition of 3D medical images. However, it will still induce false negatives as discussed in Section 1. 
%The main difference between it and our method is the strategy for obtaining positive pairs.
%Instead of using data augmentation, we utilize domain-specific structural information in volumetric medical images and consider slices of adjacent position in different patients as positive pairs.
% because they usually contain similar anatomical structures. 
%Although the global contrastive learning proposed by \cite{chaitanya2020contrastive} also considers the domain-specific information in 3D medical images, 
% The main difference is that they divided each 3D volume into several partitions and consider slices of corresponding partitions in different volume as positive pairs and those of different partitions as negative pairs, which may result in many false negative samples because the last few slices of a partition could be very similar to the first few slices of the next partition.
In contrast, the PCL method we propose uses the relative position of the slices in the volumes to decide whether they are positive pairs, thus the false negative issue can be alleviated.
In addition, the method in \cite{chaitanya2020contrastive} is only evaluated in semi-supervised setting where contrastive learning and downstream tasks are done on the same dataset.
We extend the evaluation to transfer learning to test whether the features learned by PCL on one dataset are transferable to another, and show that PCL can do better than \cite{chaitanya2020contrastive} in both settings. 

\section{Method}

\begin{figure}
\centering
  \includegraphics[width=0.96\textwidth]{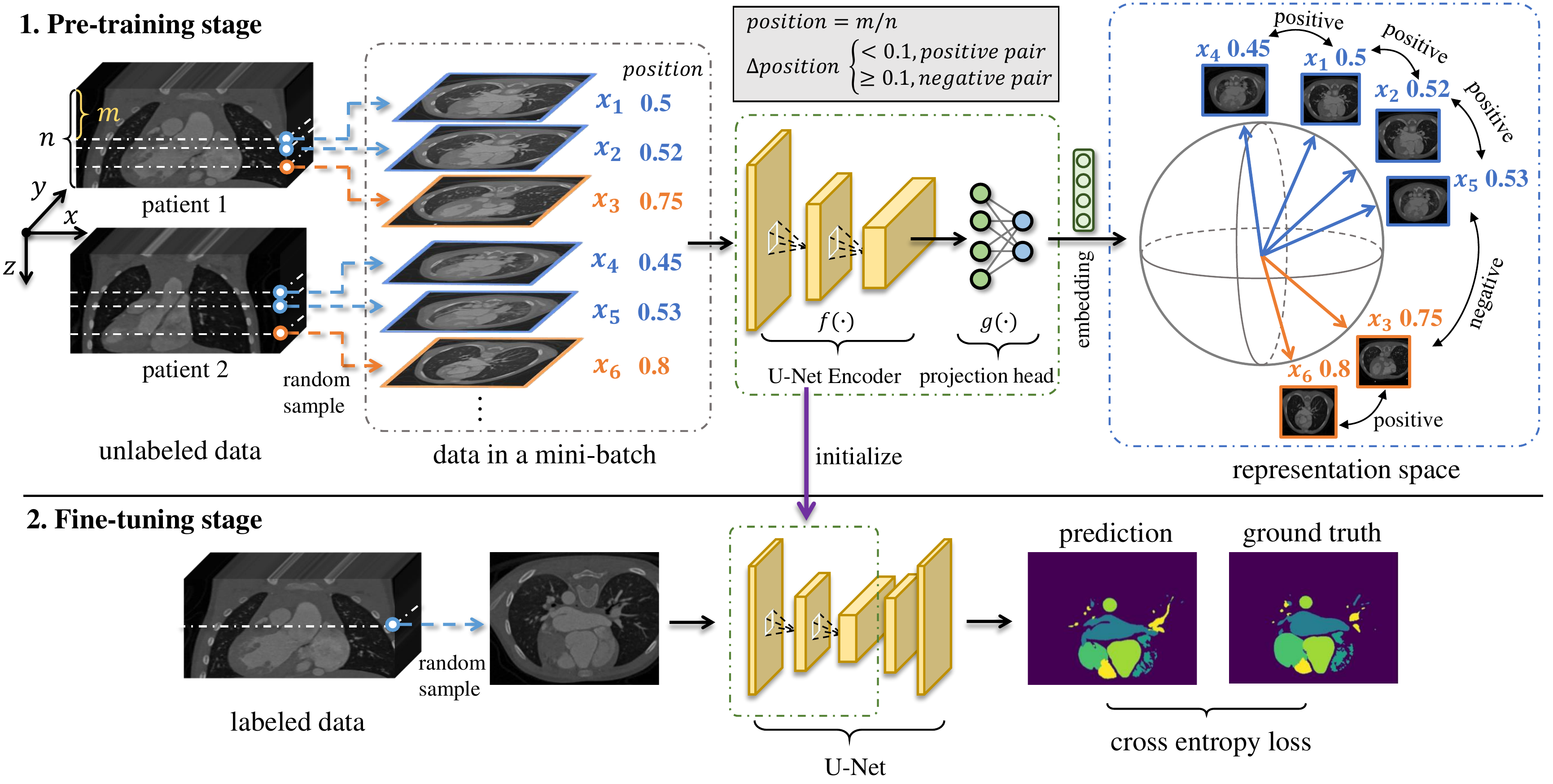}
% \vspace{-15pt}
\caption{Overview of the proposed PCL framework. In the pre-training stage, 2D slices (denoted as $x_i$) in the $xy$ plane are extracted from volumetric medical images and fitted into a U-Net encoder for representation learning.
The learned encoder is then used as initialization in the fine-tuning stage.
We use $position$ to denote the relative position of a slice along the $z$ axis in a volume.
Data pairs with small $position$ difference (e.g., $\Delta position<0.1$) are considered as positive pairs and those with large $position$ difference are considered as negative pairs.
Similar slices are marked/labeled with the same color.
}
% \vspace{-24pt}
\label{fig:framework_pcl}
\end{figure}

\subsection{Framework Overview}
In this work, for fair comparison we follow \cite{chaitanya2020contrastive} and use 2D U-Net  \cite{ronneberger2015u} to perform segmentation on 2D slices of 3D images, which has shown a remarkable success in many 3D image segmentation tasks \cite{isensee2018nnu,nemoto2020efficacy,wang2019msu,ushinsky20213d,isensee2017automatic}.
The proposed method can also be readily generalized to patch-based 3D U-Net and 3D-2D hybrid U-Net approaches.
Our PCL framework is shown in Fig.\ref{fig:framework_pcl}.
In the pre-training stage, the input of the framework is a set of 2D slices in the $xy$ plane sampled randomly from unlabeled volumetric medical images.
These slices are then propagated to a U-Net encoder $f(\cdot)$ (also known as the feature extractor) followed by a shallow multilayer perceptron (MLP) projection head $g(\cdot)$.
Let $x_i$ denote an input 2D slice. Then $h_i=f(x_i)$ is the representation learned by the encoder $f(\cdot)$ and $z_i=g(f(x_i))$ is the embedding vector.
A contrastive loss is employed on all the embeddings learned from the data in a mini-batch to perform contrastive learning.
After contrastive learning finishes, $g(\cdot)$ is thrown away and $f(\cdot)$ is used as the initialization in the standard U-Net architecture to train the network on the limited labeled dataset by supervised learning in the fine-tuning stage. 

\subsection{Leveraging Structural Information in Medical Image}
In medical images, similar anatomical structures often exist in all volumes of different patients across the dataset. In addition, we note the following two observations for volumetric medical images: 1) they have high spatial resolution along $z$ axis so that adjacent 2D slices (e.g., $x_1$ and $x_2$ in Fig. \ref{fig:framework_pcl}) inside a volume usually have similar content; 2) if the volumes of different patients are perfectly aligned, the corresponding 2D slices in different volumes (e.g., $x_2$ and $x_5$ in Fig. \ref{fig:framework_pcl}) often contain similar anatomical information.
In this paper, we utilize these two distinctive features in volumetric medical images to generate data pairs for contrastive learning.

To be specific, each 2D slice extracted from a volume is associated with a $position$ variable.
The $position$, which is between 0 and 1, represents the relative or normalized position of the slice along the $z$ axis in the volume.
Suppose $m$ is the index of the 2D slice along the $z$ axis and $n$ is the total number of slices in the $z$ axis (see Fig. \ref{fig:framework_pcl}). Then $position=m/n$. This allows the proper alignment between different volumes.
Once each 2D slice in a mini-batch is assigned with its $position$, we can use the $position$ difference to decide whether each data pair is similar or not.
If the $position$ difference of two slices is less than a threshold $t$ (e.g., 0.1 in Fig.~\ref{fig:framework_pcl}), they are likely to contain similar anatomical content and can be considered as positive pair.
Otherwise, they are negative pair.
The threshold $t$ is a hyper-parameter that is different for different medical datasets.
Note that this approach allows the positive and negative pairs to be formed on the fly instead of predefined such as in  \cite{chaitanya2020contrastive}. 
It is possible that $(x_i,x_j)$ and $(x_j,x_k)$ are positive pairs but $(x_i,x_k)$ is a negative pair.
We believe this can enforce the feature representation to be uniformly distributed on the representation hypersphere which may boost the contrastive learning performance \cite{wang2020understanding}.

As in \cite{he2020momentum,chen2020simple}, a pair of random transformations is applied for each sample in the mini-batch to help the encoder learn the spatial invariant feature of the target.
The augmentations will not change the $position$ value of the original sample, so our contrastive data pair generation strategy discussed above still works.

\subsection{Contrastive Loss Function}
Our contrastive learning loss function is based on \cite{khosla2020supervised}.
For a set of $N$ randomly sampled slices, $\{x_i\}_{i=1...N}$, the corresponding mini-batch consists of $2N$ samples after data augmentation, $\{\tilde{x}_i\}_{i=1...2N}$, in which $\tilde{x}_{2i}$ and $\tilde{x}_{2i-1}$ are two random augmentations of $x_{i}$.
$z_i$ represents the learned embedding of $\tilde{x}_i$.
Then the loss function can be defined as:
\begin{equation}
    \mathcal{L}^{PCL}=\sum_{i=1}^{2N}{\mathcal{L}_{i}^{PCL}},
\label{eq:1}
\end{equation}
\begin{equation}
    \mathcal{L}_{i}^{PCL}=-\frac{1}{|\Omega_{i}^{+}|}\sum_{j\in \Omega_{i}^{+}}{log\frac{e^{sim(z_i,z_j)/\tau}}{\sum_{k=1}^{2N}\mathbb{1}_{i\neq k}\cdot e^{sim(z_i,z_k)/\tau}}}.
\label{eq:2}
\end{equation}
where $\Omega_{i}^{+}$ is the set of indices of positive samples to $\tilde{x}_i$. $sim(\cdot,\cdot)$ is the cosine similarity function that computes the similarity between two vectors in the representation space. $\tau$ is a temperature scaling parameter.
Compared with the standard contrastive loss \cite{chen2020simple} that only has one positive pair on the numerator for any sample ${x}_i$, in Eq. 
\ref{eq:2} all positive pairs in a mini-batch (e.g., the augmented one and any of the remaining $2(N-2)$ samples whose $position$ is close to ${x}_i$) contribute to the numerator, allowing better utilization of the proposed strategy.
% The loss function can handle arbitrary numbers of positive pairs in $\Omega_{i}^{+}$.

\section{Experiments and Results}

\noindent\textbf{Datasets:}
We evaluate the performance of the proposed PCL on four publicly available medical image datasets.
\textbf{(1) The CHD dataset} is a CT dataset that consists of 68 3D cardiac images captured by a Simens biograph 64 machine \cite{xu2019whole}.
The dataset covers 14 types of congenital heart disease and the segmentation labels include seven substructures: left ventricle (LV), right ventricle (RV), left atrium (LA), right atrium (RA), myocardium (Myo), aorta (Ao) and pulmonary artery (PA).
\textbf{(2) The MMWHS dataset} was hosted in STACOM and MICCAI 2017 \cite{zhuang2016multi,zhuang2013challenges}. It consists of 20 cardiac CT and 20 MRI images and the annotations include the same seven substructures as the CHD dataset.
\textbf{(3) The ACDC dataset} was hosted in MICCAI 2017 challenge \cite{bernard2018deep}. 
The dataset has 100 patients with 3D cardiac MRI images. 
Each patient has around 15 volumes covering a full cardiac cycle, only volumes for the end-diastolic and end-systolic phase are labeled by an expert.
The segmentation labels include three substructures: LV, RV, and Myo.
\textbf{(4) The HVSMR dataset} was hosted in MICCAI 2016 challenge \cite{pace2015interactive}. It has 10 3D cardiac MRI images captured in an axial view on a 1.5T scanner. Manual annotations of blood pool and Myo are provided.

\noindent\textbf{Preprocessing:}
Following the work of \cite{chaitanya2020contrastive}, we first normalize the intensity of each 3D volume $x$ to [$x_1, x_{99}$], where $x_p$ is the $p$-th intensity percentile in $x$.
Then all 2D slices and the corresponding annotations are resampled to a fixed spatial resolution $f_{r}$ and padded to a fixed image size $f_{s}$ with 0.
We do not apply cropping because it may remove important structure information in the original slice.
The $f_{r}$ and $f_{s}$ for each dataset are defined as follows (1) CHD dataset: $f_{r}=1.0\times 1.0 mm^2$ and $f_{s}=512\times 512$, (2) MMWHS dataset: $f_{r}=1.0\times 1.0 mm^2$ and $f_{s}=256\times 256$, (3) ACDC dataset: $f_{r}=1.25\times 1.25 mm^2$ and $f_{s}=352\times 352$,
(4) HVSMR dataset: $f_{r}=0.7\times 0.7 mm^2$ and $f_{s}=352\times 352$.
No additional alignment technique is used for CHD and ACDC datasets because they are already roughly aligned as they are acquired.

\subsection{Semi-supervised Learning}
In this section, we test whether the proposed PCL can improve the performance in semi-supervised learning where contrastive learning and down-stream supervised learning (with limited annotation) are done on the same dataset. 
%Specifically, the contrastive learning is performed on CHD and ACDC without labels, and then the pre-trained models are fine tuned on the same dataset with limited labels.

\noindent\textbf{Setup:}
We employ our PCL to pre-train a U-Net encoder on the whole CHD  and ACDC, respectively, without using any human label.
Note that for ACDC, each patient has more than 10 volumes covering a full cardiac cycle, only two of which have annotations. Since we do not need labels anyway, we use all the volumes from 100 patients for pre-training.
Then the pre-trained model is used as the initialization to fine-tune a U-Net segmentation network with a small number of labeled samples on the same dataset.
5-fold cross-validation is used to evaluate the segmentation performance.
Specifically, for each cross-validation fold on CHD, 
% 51 patients \textit{can be used} for training and the rest 17 patients are used for evaluation in each fold for the fine-tuning stage.
We randomly sample $M$ patients from the 51 patients for fine-tuning, as if we only have the labels for these patients, and evaluate the results on the remaining 17 patients.
We experiment with different values of $M$ (e.g., 2, 6 and 10) to assess the influence of training set size in the fine-tuning stage on the contrastive learning performance.
The same training strategy is also used for ACDC.
We choose the threshold $t$ to be 0.1 and 0.35 for CHD and ACDC because they have the best performance according to our experiment.
The influence of thresholds on accuracy will be discussed in the supplementary.
Data augmentations, including translation, rotation, and scale, are used in both the pre-training and fine-tuning stages.
The pre-training is done on two NVIDIA GeForce GTX 1080 GPUs with 200 epochs.
SGD is used as the optimizer and the learning rate is set to 0.1. We use cosine learning rate scheduler, batch size is set to 32. Temperature $\tau$ is set to 0.1 as in \cite{he2020momentum,chen2020simple}.
In the fine-tuning stage, we train the U-Net with cross-entropy loss for 100 epochs. The batch size is set to 5 and the learning rate is $5\times e^{-5}$. Adam optimizer and cosine scheduler are used.

\noindent\textbf{Baselines:} We compare the performance of PCL with a random approach that does not use any pre-training as well as the following state-of-the-art baselines, all of which use the same unlabeled dataset in the pre-training and labeled dataset in the fine-tuning as PCL: 
(1) Rotation \cite{gidaris2018unsupervised}: a pretext-based method that uses image rotation prediction to pre-train the encoder; 
(2) PIRL \cite{misra2020self}: adopted from a contrastive learning scheme for natural image classification, which uses contrastive loss to learn pretext-invariant representations.
(3) SimCLR \cite{chen2020simple}: adopted from another contrastive learning scheme for natural image classification, which constructs positive pairs for each sample only using two random augmentations; 
(4) GCL \cite{chaitanya2020contrastive}: a contrastive learning scheme for 3D medical image segmentation which divides each volume into four partitions so that slices belonging to the same partition in different volumes are considered as positive pairs.

\begin{table}[t]
\centering
\caption{Comparison of the proposed PCL method with baseline methods on CHD and ACDC. $M$ is the number of patients used in the fine-tuning process. Results are reported in the form of mean(standard deviation) on 5-fold cross-validation. PCL provides better results than the baselines for all values of $M$. 
% The improvement is more significant with smaller training data size during fine-tuning.
}
\resizebox{1.0\columnwidth}{!}{
\begin{tabular}{lccccccc}
\toprule
\multicolumn{8}{c}{CHD (68 patients in total)} \\
Method & $M$=2 & $M$=6 & $M$=10 & $M$=15 & $M$=20 & $M$=30 & $M$=51 \\ \hline 
Random & 0.184(.06) & 0.508(.06) & 0.584(.05) & 0.627(.05) & 0.658(.04) & 0.693(.04) & 0.754(.02) \\ 
Rotation~\cite{gidaris2018unsupervised} & 0.171(.06) & 0.488(.07) & 0.575(.04) & 0.625(.04) & 0.651(.04) & 0.691(.04) & 0.749(.03) \\
PIRL~\cite{misra2020self} & 0.196(.07) & 0.504(.08) & 0.617(.05) & 0.658(.03) & 0.674(.04) & 0.714(.04) & 0.761(.03) \\
SimCLR~\cite{chen2020simple} & 0.192(.06) & 0.515(.06) & 0.599(.06) & 0.631(.05) & 0.666(.05) & 0.699(.05) & 0.756(.03) \\
GCL~\cite{chaitanya2020contrastive} & 0.255(.10) & 0.564(.04) & 0.646(.03) & 0.669(.04) & 0.697(.04) & 0.725(.04) & 0.766(.03)        \\
PCL & \textbf{0.356(.08)} & \textbf{0.600(.06)} & \textbf{0.661(.05)} & \textbf{0.686(.05)} & \textbf{0.716(.04)} & \textbf{0.735(.05)} & \textbf{0.774(.03)} \\
\bottomrule
% \multicolumn{2}{|c|}{Average Score} & 3.61 & 4.125 \\ \hline
\end{tabular}
}

% \vspace*{0.2 cm}

\resizebox{1.0\columnwidth}{!}{
\begin{tabular}{lccccccc}
\toprule
\multicolumn{8}{c}{ACDC (100 patients in total)} \\
Method & $M$=2 & $M$=6 & $M$=10 & $M$=15 & $M$=20 & $M$=30 & $M$=80 \\ \hline 
Random & 0.588(.07) & 0.782(.03) & 0.840(.03) & 0.876(.01) & 0.894(.01) & 0.909(.01) & 0.928(.00) \\ 
Rotation~\cite{gidaris2018unsupervised} & 0.572(.08) & 0.809(.03) & 0.868(.02) & 0.886(.01) & 0.898(.01) & 0.910(.01) & 0.925(.00) \\
PIRL~\cite{misra2020self} & 0.492(.03) & 0.823(.04) & 0.865(.01) & 0.880(.02) & 0.896(.02) & 0.912(.01) & 0.927(.00) \\
SimCLR~\cite{chen2020simple} & 0.352(.06) & 0.725(.08) & 0.824(.04) & 0.869(.02) & 0.894(.01) & 0.913(.01) & 0.927(.00) \\
GCL~\cite{chaitanya2020contrastive} & 0.636(.05) & 0.803(.04) & 0.872(.01) & 0.891(.01) & 0.902(.01) & 0.913(.01) & 0.927(.01) \\
PCL & \textbf{0.671(.06)} & \textbf{0.850(.01)} & \textbf{0.885(.01)} & \textbf{0.904(.01)} & \textbf{0.909(.01)} & \textbf{0.919(.00)} & \textbf{0.929(.00)} \\
\bottomrule
% \multicolumn{2}{|c|}{Average Score} & 3.61 & 4.125 \\ \hline
\end{tabular}
}
% \vspace{-15pt}
\label{fig:result_semi_supervised}
\end{table}

% \begin{figure}[t]
% \centering
%   \includegraphics[width=\textwidth]{figs/semi_supervised_result.pdf}
% \caption{Comparison of the proposed PCL method with other baseline pre-training methods on (a) CHD dataset and (b) ACDC dataset. The proposed PCL provides better results than the baselines on both datasets in all settings. The improvement is more significant with smaller training samples during fine-tuning.}
% \label{fig:result_semi_supervised}
% \vspace{-0.1in}
% \end{figure}

\noindent\textbf{Results and Analysis:}
The results of the comparative study on both CHD and ACDC are shown in Table \ref{fig:result_semi_supervised}.
We report the averaging Dice of 5-fold cross-validation results.
From the table, we have the following observations. 
(1) Comparing PCL and GCL with other baselines, we can see that the performance improves significantly ($\Delta Dice>0.1$) in many settings for both CHD and ACDC, suggesting that by leveraging domain-specific structural information in volumetric medical images, the encoder can learn better task-related representation for segmentation.
(2) The performance improvement of PCL and GCL are especially high when a very small number of training samples are used (e.g., 2 and 4). The gains become lesser when the number of training samples increases.
This is because with more training samples, the information difference between the training set for fine-tuning and the training set for contrastive learning becomes small and the fine-tuning performance saturates.
(3) SimCLR performs worse than Random on ACDC.
This suggests that only using data augmentations to generate contrastive data pairs may lead to a large false negative rate for datasets like ACDC where the volumes have small $z$ dimensions (around 10).
(4) PCL performs better than GCL in all settings. The improvement in Dice can be up to $0.04$.
This shows that using the relative $position$ difference instead of a hard partition strategy can better utilize the structural information in medical images and reduce false negatives to improve contrastive learning performance.
% pick up some special point.

\subsection{Transfer Learning}

% \begin{figure}[t]
% \centering
%   \includegraphics[width=\textwidth]{figs/semi_supervised_result.pdf}
% \caption{Transfer learning comparison of the proposed PCL method with baselines on (a) MMWHS dataset and (b) HVSMR dataset. The proposed PCL shows better transfer performance than the baselines on both datasets.}
% \label{fig:result_transfer_learning}
% \end{figure}

\begin{table}[t]
\centering
\caption{Transfer learning comparison of the proposed PCL method with the baselines. Except for Random, all the methods are pre-trained on CHD and ACDC without labels and fine-tuned on MMWHS and HVSMR respectively. }
% PCL shows better transfer performance than the baselines on both datasets.}
\resizebox{1.0\columnwidth}{!}{
\begin{tabular}{lcccccc}
\toprule
\multicolumn{7}{c}{CHD transferring to MMWHS (20 patients in total)} \\
Method & $M$=2 & $M$=4 & $M$=6 & $M$=8 & $M$=10 & $M$=16 \\ \hline 
Random & 0.232(.14) & 0.661(.10) & 0.732(.07) & 0.769(.06) & 0.808(.05) & 0.834(.05)  \\ 
Rotation~\cite{gidaris2018unsupervised} & 0.247(.16) & 0.659(.13) & 0.751(.07) & 0.768(.07) & 0.803(.06) & 0.850(.04) \\
PIRL~\cite{misra2020self} & 0.251(.10) & 0.670(.11) & 0.755(.07) & 0.774(.06) & 0.821(.05) & 0.851(.04) \\
SimCLR~\cite{chen2020simple} & 0.269(.17) & 0.683(.10) & 0.751(.07) & 0.783(.06) & 0.818(.05) & 0.850(.04) \\
GCL~\cite{chaitanya2020contrastive} & 0.262(.11) & 0.703(.07) & 0.768(.05) & 0.805(.04) & 0.820(.04) & 0.851(.03) \\
PCL & \textbf{0.339(.15)} & \textbf{0.748(.08)} & \textbf{0.792(.05)} & \textbf{0.820(.05)} & \textbf{0.840(.04)} & \textbf{0.869(.03)} \\
\bottomrule
% \multicolumn{2}{|c|}{Average Score} & 3.61 & 4.125 \\ \hline
\end{tabular}
}

% \vspace*{0.2 cm}

% \resizebox{0.65\columnwidth}{!}{
\begin{tabular}{lcccc}
\toprule
\multicolumn{5}{c}{ACDC transferring to HVSMR (10 patients in total)} \\
Method & $M$=2 & $M$=4 & $M$=6 & $M$=8 \\ \hline 
Random & 0.742(.06) & 0.813(.05) & 0.842(.03) & 0.842(.04) \\ 
Rotation~\cite{gidaris2018unsupervised} & 0.737(.07) & 0.816(.06) & 0.845(.03) & 0.844(.03)  \\
PIRL~\cite{misra2020self} & 0.740(.05) & 0.826(.04) & 0.849(.03) & 0.846(.03) \\
SimCLR~\cite{chen2020simple} & 0.700(.07) & 0.779(.05) & 0.808(.04) & 0.815(.04) \\
GCL~\cite{chaitanya2020contrastive} & 0.770(.05) & 0.818(.05) & 0.842(.03) & 0.843(.03) \\
PCL & \textbf{0.781(.05)} & \textbf{0.832(.05)} & \textbf{0.857(.03)} & \textbf{0.857(.03)} \\
\bottomrule
% \multicolumn{2}{|c|}{Average Score} & 3.61 & 4.125 \\ \hline
\end{tabular}
% }
% \vspace*{-0.3 cm}
\label{fig:result_transfer_learning}
\end{table}

To assess whether the learned representations by PCL are transferrable, we use the encoder pre-trained on CHD and ACDC without labels as the initialization of a U-Net to fine-tune on MMWHS and HVSMR datasets respectively.
The experiment setup and baselines are the same as in Section 4.1.
Table \ref{fig:result_transfer_learning} shows the comparison results.
It can be seen that the proposed PCL framework outperforms all baselines on both datasets.
The overall improvement on HVSMR is relatively smaller than MMWHS. This is because MMWHS is very similar to CHD which makes the features learned on CHD more helpful on MMWHS.
On the other hand, ACDC and HVSMR are different in terms of acquisition view and image resolution, which limits the transfer learning performance.
Visualization of the segmentation results on all datasets is shown in the supplementary.

\section{Conclusion}

In this paper, we propose a novel PCL framework for representation learning in volumetric medical images. 
The framework can effectively eliminate false negative pairs in existing contrastive learning methods for medical image segmentation. 
Experimental results on four 3D medical image datasets show that PCL significantly improves the segmentation performance in both semi-supervised setting and transfer learning setting.

\noindent\textbf{Acknowledgements.} This work is partially supported by NSF award IIS-2039538.

\bibliographystyle{splncs04}
\bibliography{reference.bib}

\end{document}

\title{Supplementary: Positional Contrastive Learning for Volumetric Medical Image Segmentation}

\author{}
\institute{}

\maketitle             

\begin{table}[ht]
\centering
\caption{Ablation study. The influence of threshold on contrastive learning accuracy. Results are reported in the form of mean(standard deviation) on 5-fold cross-validation. $M$ is the number of patients used in the fine-tuning stage. The average number of positive pairs at each threshold is reported. The results show that there exists an optimal threshold with the best contrastive learning performance for different datasets. When the threshold is too small, the false negative rate will increase. When the threshold is too large, the false positive rate will increase. Both situations will induce performance degradation.}
\begin{tabular}{cccccc}
\toprule
\multicolumn{5}{c}{CHD} \\
Threshold & $\#$positive pairs &  $M$=10 & $M$=15 & $M$=20 & $M$=30 \\ \hline 
0.05 & 6.98 & 0.651(.06) & 0.681(.05) & 0.710(.06) & 0.733(.05) \\
0.10 & 12.69 & \textbf{0.661(.05)} & \textbf{0.686(.05)} & \textbf{0.716(.04)} & \textbf{0.735(.05)} \\
0.15 & 18.23 & 0.658(.05) & 0.684(.06) & \textbf{0.716(.04)} & 0.734(.06) \\
0.20 & 23.29 & 0.655(.05) & 0.680(.07) & 0.711(.05) & 0.732(.06) \\ 
\bottomrule
\end{tabular}

\vspace*{0.2 cm}

\begin{tabular}{cccccc}
\toprule
\multicolumn{5}{c}{ACDC} \\
Threshold & $\#$positive pairs & $M$=10 & $M$=15 & $M$=20 & $M$=30 \\ \hline 
0.20 & 23.66 & 0.877(.02) & \textbf{0.904(.01)} & 0.910(.01) & 0.915(.01) \\
0.25 & 27.66 & 0.879(.01) & 0.901(.01) & 0.912(.01) & 0.917(.01) \\
0.30 & 31.93 & 0.870(.02) & \textbf{0.904(.00)} & 0.912(.00) & 0.918(.00) \\
0.35 & 36.69 & \textbf{0.885(.01)} & \textbf{0.904(.01)} & \textbf{0.913(.01)} & \textbf{0.919(.00)} \\
0.40 & 40.32 & 0.877(.02) & 0.897(.01) & 0.905(.01) & 0.918(.01) \\
0.50 & 46.82 & 0.883(.01) & 0.900(.01) & 0.910(.01) & \textbf{0.919(.00)} \\ \bottomrule
\end{tabular}
\label{fig:ablation_threshold}
\end{table}

\begin{figure}[ht]
\centering
  \includegraphics[width=\textwidth]{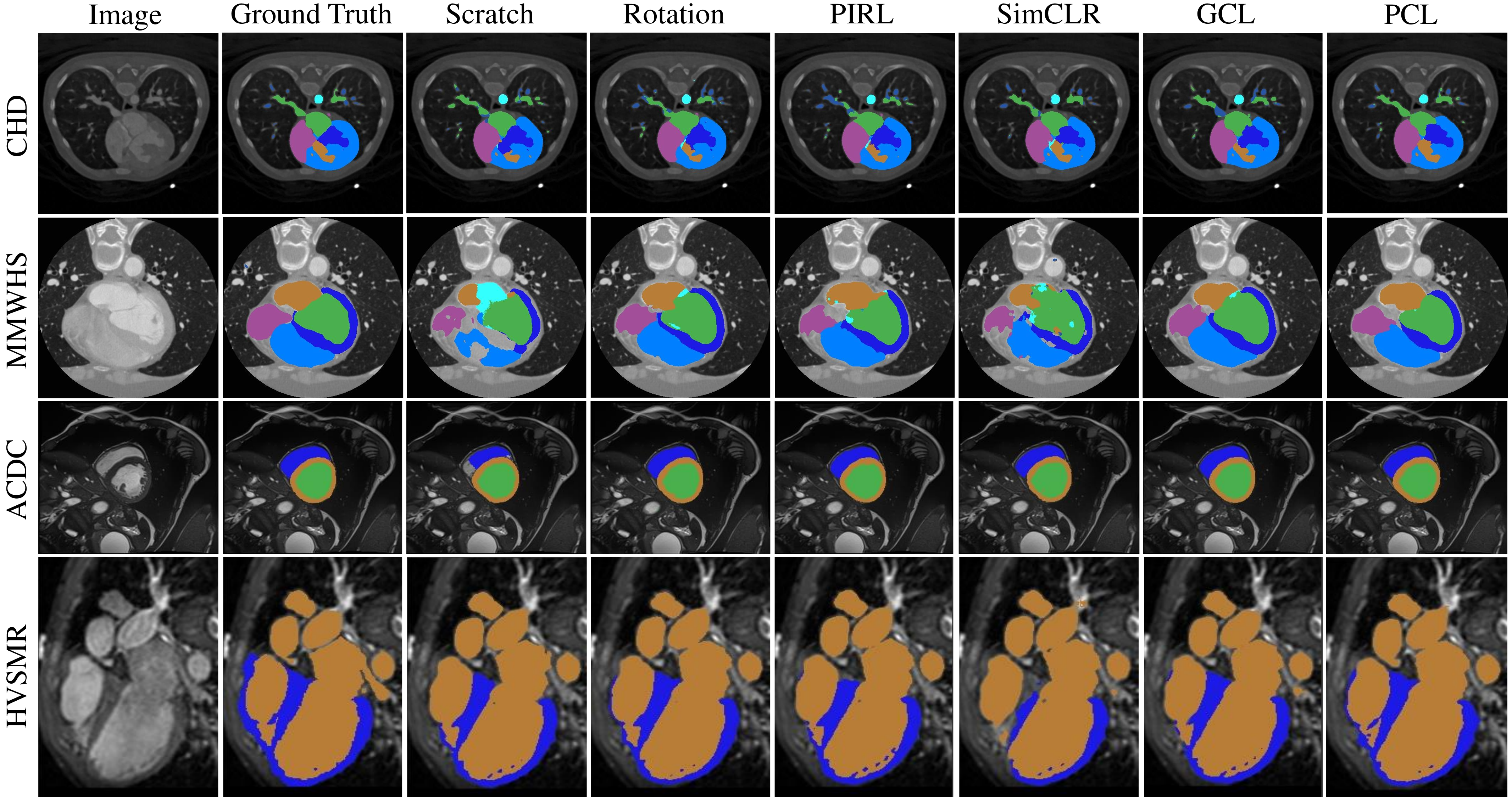}
% \vspace{-15pt}
\caption{Visualization of segmentation results on all datasets. The results of CHD and ACDC are generated from the fine-tuned model when $M=10$. The results of MMWHS and HVSMR are generated from the fine-tuned model when $M=6$. $M$ is the number of patients used for fine-tuning.}
\label{fig:framework_scl}
\end{figure}